\documentclass[letterpaper]{article} 
\usepackage{aaai24}  
\usepackage{times}  
\usepackage{helvet}  
\usepackage{courier}  
\usepackage[hyphens]{url}  
\usepackage{graphicx} 
\usepackage{natbib}  
\usepackage{caption} 
\usepackage{algorithm}
\usepackage{algorithmic}

\usepackage{multirow}
\usepackage{threeparttable}
\usepackage{amsmath} 

\usepackage{newfloat}
\usepackage{listings}
\usepackage{booktabs}
\usepackage{bbding}
\usepackage[table]{xcolor}
\usepackage{stix}
\DeclareCaptionStyle{ruled}{labelfont=normalfont,labelsep=colon,strut=off} 
\floatstyle{ruled}
\newfloat{listing}{tb}{lst}{}
\floatname{listing}{Listing}
\title{Mining Fine-Grained Image-Text Alignment for Zero-Shot Captioning \\via Text-Only Training}

\author{
   Longtian Qiu\textsuperscript{\rm1}\equalcontrib, Shan Ning\textsuperscript{\rm1}\equalcontrib, Xuming He\textsuperscript{\rm1,2}\\
}
\affiliations{
    ShanghaiTech University, Shanghai, China\textsuperscript{\rm1} \\ Shanghai Engineering Research Center of Intelligent Vision and Imaging\textsuperscript{\rm2}\\
    {\tt\small {\{qiult, ningshan2022, hexm\}@shanghaitech.edu.cn}}

}

\usepackage{bibentry}

\begin{document}

\maketitle

\begin{abstract}


Image captioning aims at generating descriptive and meaningful textual descriptions of images, enabling a broad range of vision-language applications. Prior works have demonstrated that harnessing the power of Contrastive Image Language Pre-training (CLIP) offers a promising approach to achieving zero-shot captioning, eliminating the need for expensive caption annotations. However, the widely observed modality gap in the latent space of CLIP harms the performance of zero-shot captioning by breaking the alignment between paired image-text features. To address this issue, we conduct an analysis on the CLIP latent space which leads to two findings. Firstly, we observe that the CLIP's visual feature of image subregions can achieve closer proximity to the paired caption due to the inherent information loss in text descriptions. In addition, we show that the modality gap between a paired image-text can be empirically modeled as a zero-mean Gaussian distribution. Motivated by the findings, we propose a novel zero-shot image captioning framework with text-only training to reduce the modality gap. In particular, we introduce a subregion feature aggregation to leverage local region information, which produces a compact visual representation for matching text representation. Moreover, we incorporate a noise injection and CLIP reranking strategy to boost captioning performance. We also extend our framework to build a zero-shot VQA pipeline, demonstrating its generality. Through extensive experiments on common captioning and VQA datasets such as MSCOCO, Flickr30k and VQAV2, we show that our method achieves remarkable performance improvements. Code is available at https://github.com/Artanic30/MacCap.

\end{abstract}
\section{Introduction}
Image captioning is a fundamental task in vision-language understanding that involves generating natural language descriptions for a given image. It plays a critical role in facilitating more complex vision-language tasks, such as visual question answering \cite{Agrawal2015VQAVQ,gqa,okvqa} and visual dialog \cite{Das2016VisualD,Niu2018RecursiveVA,llava}.
The mainstream image captioning methods \cite{conimgcap4,conimgcap1,conimgcap3,conimgcap2} require expensive human annotation of image-text pairs for training neural network models in an end-to-end manner. Recent developments in Contrastive Image Language Pre-training (CLIP) \cite{clip} have enabled researchers to explore a new paradigm, zero-shot image captioning, through text-only training. In particular, CLIP learns a multi-modal embedding space where semantically related images and text are encoded into features with close proximity. As such, if a model learns to map the CLIP text features to their corresponding texts, it is feasible to generate image captions from the CLIP image features without needing supervision from caption annotations.


One main advantage of this zero-shot captioning paradigm is that it enables a Large Language Model (LLM) \cite{gpt3, Zhang2022OPTOP} with image captioning capabilities using only text data and affordable computational resources. Despite the impressive performance achieved by recent powerful multimodal models \cite{miniGPT4,llava}, they typically require large-scale, high-quality human-annotated data and expensive computational resources for fine-tuning an LLM. Zero-shot captioning methods can significantly reduce such costs, which is particularly important in situations of data scarcity and limited resources. Moreover, recent work \cite{Guo2022FromIT, Changpinyo2022AllYM,Tiong2022PlugandPlayVZ} demonstrates that other vision-language tasks, such as VQA, can be addressed by LLMs and image captions. Consequently, the paradigm of zero-shot captioning has the potential to pave the way to solving complex vision-language tasks with LLMs through efficient text-only training.

A critical challenge in zero-shot image captioning through text-only training is to mitigate a widely observed phenomenon known as the \textit{modality gap}. While the features of paired texts and images are close in the CLIP embedding space, there remains a gap between them \cite{MindGap}. This gap often results in inaccurate mappings from the image embeddings to the text ones. Consequently, without fine-tuning with paired data, it significantly impairs the performance of zero-shot image captioning.
Several works have attempted to address the modality gap in zero-shot image captioning, relying mainly on two strategies: (1) The first strategy leverages a memory bank from training text data to project visual embeddings into the text embedding space \cite{DeCap}. However, this projection prevents it from representing any semantic content outside the distribution of the memory bank features and introduces extra inference costs; (2) The second approach injects noise during training to encourage the visual embeddings to be included inside the semantic neighborhood of the corresponding text embeddings \cite{CapDec}. Nonetheless, the noise injection tends to diffuse the distribution of visual inputs at the cost of weakening the semantic correlation between paired images and text embeddings. 



To tackle these challenges, we first conduct a thorough analysis of the CLIP feature space, leading to two key observations. First, most text descriptions are unable to fully capture the content of their paired images. However, we empirically find that the visual embedding of certain local regions of an image, named image subregions, have closer proximity to the text embedding of the paired caption. Integrating such image subregions with the global image representation generates a tighter alignment between image and text. Additionally, we analyze the distribution of the gap between the CLIP features of image or subregion-text pairs and find that it closely resembles a zero-mean Gaussian distribution.


Based on our findings, we propose a novel zero-shot image captioning framework, named \textit{\textbf{M}ining Fine-Grained Image-Text \textbf{A}lignment in \textbf{C}LIP for \textbf{Cap}tioning} (MacCap), to address the aforementioned challenges. In this framework, we introduce a region-aware cross-modal representation based on CLIP and an effective unimodal training strategy for an LLM-based caption generator. Our cross-modal representation maps an input image into the language space of LLMs and consists of two main components. First, we design a \textit{sub-region feature aggregation} module to fuse both global and subregion-level CLIP image features, resulting in a smaller gap between the corresponding CLIP text embedding. Next, we introduce a learnable adaptor-decoder to transform the CLIP representation into the LLM's language space.
To train our model with text-only data, we develop a robust procedure to learn a projection from the CLIP embedding space to a language representation, enabling the LLM to reconstruct captions. Specifically, our learning procedure first injects noise into our region-aware CLIP-text representation, mimicking the modality gap between image and text features. This is followed by a multiple sampling and filtering step that leverages the CLIP knowledge to improve the quality of the captioning.
In addition to the image captioning task, we further extend our framework to build a zero-shot VQA pipeline, demonstrating the generality of our cross-modal representation for more complex vision-language tasks.


We evaluate our framework on several widely-adopted image captioning benchmarks, such as MSCOCO \cite{mscoco} and Flickr30k \cite{Flickr30k}, as well as a standard VQA benchmark, VQAV2 \cite{vqav2}. Our extensive experiments cover multiple vision-language tasks, including zero-shot in-domain image captioning, zero-shot cross-domain image, and zero-shot VQA. The results not only demonstrate the superiority of our methods but also validate our findings on the CLIP embedding space.




\section{Related Work}

\paragraph{Zero-shot Image Captioning}
Zero-shot image captioning is an emerging task setting of image captioning, where captions are generated without relying on training with annotated image data. While some approaches \cite{Changpinyo,Wang_Yu_Yu_Dai_Tsvetkov_Cao_2021,flamingo} exploit large noisy image-text datasets, demanding high data and computational resources, an alternative is to leverage pre-trained large models, which is more suitable for low-data scenarios.

The use of pre-trained multi-modality models has enabled progress in text-only training for image captioning, which has demonstrated promising results. CapDec \cite{CapDec} utilizes CLIP embeddings and employs a noise injection training strategy for text-only training. Similarly, DeCap \cite{DeCap} employs a memory bank to project visual features into the text modality. Furthermore, methods like MAGIC \cite{MAGIC} and ZeroCap \cite{ZeroCap} achieve zero-shot captioning without a typical training stage, with MAGIC introducing a CLIP-based score to guide language model generation and ZeroCap employing iterative optimization during inference.

\paragraph{Vision-language Models}

Recent advancements \cite{clip,align,glip,blip,glip2} in Vision-Language Models (VLM) have led to significant progress in various downstream tasks \cite{styleclip,maskclip,pointclip,denclip,calip,hoiclip}. Extensive research efforts \cite{Wang2020UnderstandingCR,Wang2020UnderstandingTB,clipbagofwords,MindGap} have analyzed the multi-modal embedding space learned through contrastive training. Recently, a geometric phenomenon known as the \textit{modality gap} has been identified in \cite{MindGap}. This gap arises from misalignment between text and image embeddings of CLIP, impacting their shared representation. The \textit{modality gap} is attributed to the optimization process of contrastive learning and random initialization of different encoders. Addressing this \textit{modality gap} is crucial for enhancing zero-shot capabilities, especially in scenarios with limited fine-tuning opportunities.

\section{CLIP Embedding Space Analysis}
In this section, we present a detailed analysis of the CLIP embedding space. CLIP offers a pre-trained joint embedding space, enabling zero-shot vision-language tasks. However, there is still a \textit{modality gap} present in the CLIP embedding space \cite{MindGap}. This gap refers to a geometric property where image and text embeddings occupy distinct regions within the embedding space. We further analyze and demonstrate that the modality gap arises from an inherent ambiguity in matching visual and linguistic embeddings. Empirically, we show that integrating subregion image information with global information can reduce the gap between linguistic and visual representations. Additionally, we explore the disparity between CLIP's text and image representations, conducting an empirical study that reveals the difference follows a Gaussian distribution. These findings inspire our subsequent method design.
  

\begin{figure}[t!]
  \centering
  \includegraphics[width=0.35\textwidth]{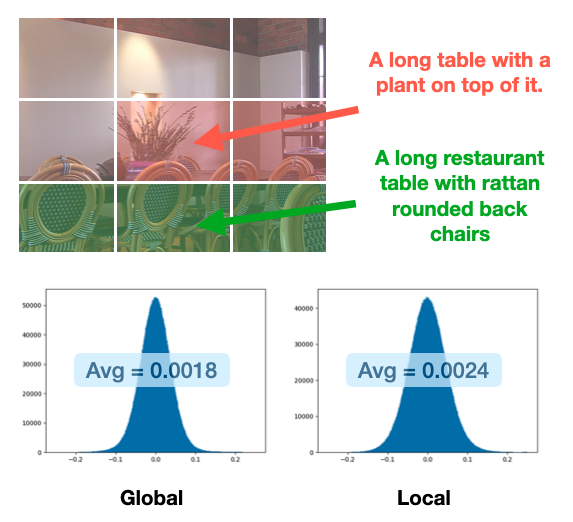}
   \caption{The upper half of this figure is an example of the misalignment in paired image and text description. The lower half of this figure is the distribution of modality gap between text representation and global / local image representation respectively.}
    \label{figure:ex}
\end{figure}

\begin{table}[t]
\small
\centering
\tabcolsep=2pt
\begin{tabular}{c|ccc}
\toprule
\toprule
\multirow{2}{*}{} & \multicolumn{3}{l}{Pair Cosine Similarity} \\
                         & Mean        & Max        & Min        \\ \hline
 Global representation           & 0.330       & 0.446      & 0.228      \\ \hline

Mix representation             & 0.352       & 0.422      & 0.242      \\ \toprule
\end{tabular}
\caption{In this table, we show the mean, max and min value of similarity between text feature and global/mix feature. We find that after adding subregion information, the mean, max and min value all increase. This observation shows that introducing subregion image information benefit the alleviation of modality gap.}
\label{table:subregion}
\end{table}

\subsection{Modality Gap of CLIP Representations}

As demonstrated in \cite{MindGap}, the modality gap phenomenon is primarily caused by the presence of mismatched text-image data during pre-training, further exacerbated by the contrastive learning objective employed in CLIP. However, we observe that even correctly paired image-text may exhibit different semantic contents due to several reasons, including 1) incomplete image description by text; 2) ambiguity in text interpretation; 3) diverse valid descriptions with varying focus on image subregions. An example is illustrated in Figure~\ref{figure:ex}.
Consequently, we observe that certain subregions of an image typically exhibit a smaller modality gap with a specific text description, as illustrated in Figure~\ref{figure:ex}. To validate this characteristic, we conduct a statistical analysis on the MSCOCO validation set. Specifically, we calculate the cosine similarity between the visual and text features, where the visual features are obtained from the global CLS token or the last layer of the ViT encoder representing the subregions. The results indicate that in $\mathbf{33\%}$ of cases, one of the subregion features has a higher similarity than the global one.



Motivated by the above observation, we propose integrating the global and subregion representations by adding them together. We evaluate the similarity scores between text features and their corresponding subregion-augmented image representations. As illustrated in Table~\ref{table:subregion}, our augmented image embeddings achieve higher similarity scores when compared with corresponding text embeddings without any fine-tuning, effectively reducing the modality gap.

\subsection{Distribution of Modality Gap}
We further investigate the distribution of the modality gap through an empirical study on the MSCOCO validation set. Specifically, for each image with its caption, we compute the difference between the image and text embeddings. We calculate these differences separately for the global image feature and the subregion feature. Given these differences, we pool all the feature dimensions together and compute a histogram and mean of the dimension-wise differences between the two modalities. As depicted in the lower half of Figure~\ref{figure:ex}, we observe that the mean of the gap distribution is close to zero, and the overall distribution resembles a Gaussian distribution for both global and subregion representations. Based on this observation, we adopt a unimodal learning strategy that involves Gaussian noise injection with the text features. This strategy allows us to mimic the image features in the cross-modal inference stage. For a more formal description of our analysis, please refer to the supplementary material.

\begin{figure*}[t!]
  \centering
  \includegraphics[width=0.7\textwidth]{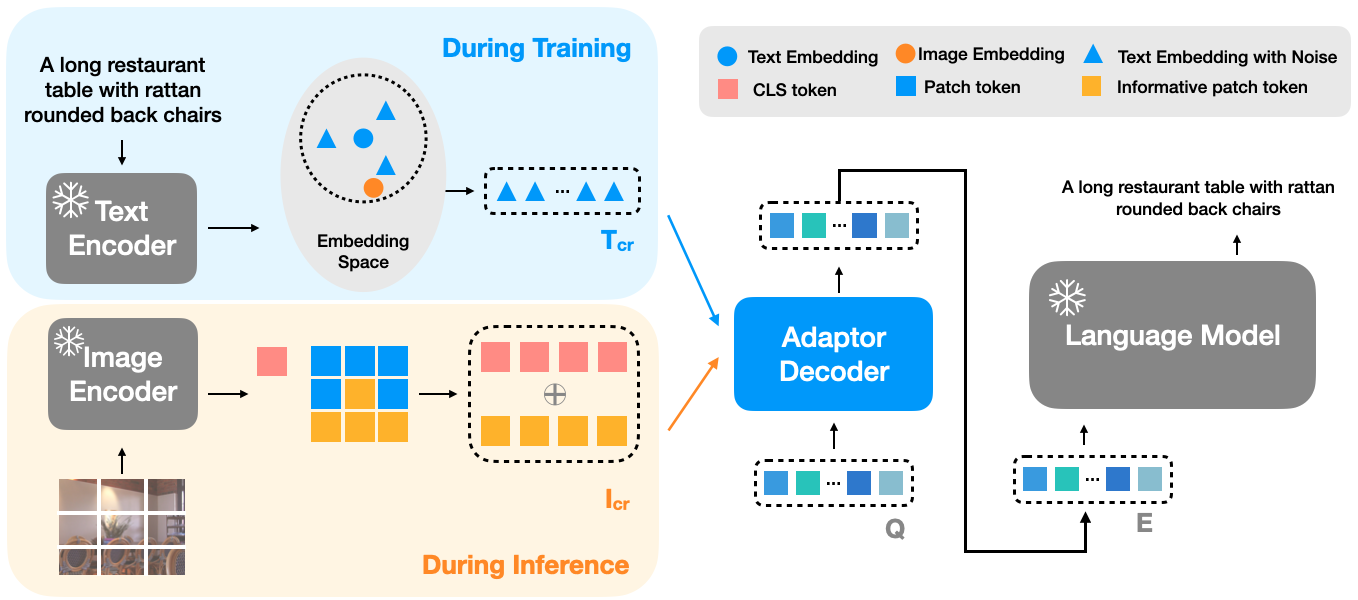}
   \caption{\textbf{An overview of MacCap pipeline.} MacCap learns to generate text based on region noise injected CLIP text feature in text reconstruction training. During inference, MacCap can generate caption without paired data in training. The CLIP and language model are kept frozen in both stages. }
    \label{figure:pipeline}
\end{figure*}

\section{Methodology}

\subsection{Method Overview}
\label{method:definition}
In zero-shot image captioning, the image captioning model is trained with only caption text data. This is possible since CLIP learned a joint space where semantically-related image feature $I_c$ and text feature $T_c$ have closer proximity. By training the model to generate captions conditioned on their CLIP text feature, the model becomes capable of generating captions based on the CLIP image feature without any supervision from paired caption data.

Specifically, we have a caption text corpus $T=\{ t^i | i\in \mathbb{N} \}$ and three network modules, contrastive vision language model CLIP with parameter $\theta_{c}$, pre-trained large language model with parameter $\theta_{l}$ and a learnable adaptor decoder with parameter $\theta$. In text reconstruction training, the adaptor module converts text $t_i$'s CLIP text feature $T_c^{i} \in R^{D}$ to a prefix embedding $E^{i} \in R^{N_q \times D_l}$, where $N_q$ is the length of prefix embedding, $D_l$ is the dimensionality of language model and $D$ is the dimensionality of CLIP feature. The language model generates text $t_i$ based on the prefix embedding $E^{i}$. During training, we freeze the parameter of CLIP $\theta_{c}$ and language model $\theta_{l}$, which makes the adaptor decoder with parameter $\theta$ a plug-and-play module to achieve zero-shot captioning. 
We can formulate the process as follows:
\begin{align}
     \underset{\theta}{\mathrm{maximize}} \quad p(t^i|T_c^{i}, \theta_{c}, \theta_{l})
\end{align}
In inference, CLIP extracts image feature $I_c \in R^{D}$ for an image. The adaptor decoder converts it to prefix embedding and the language model generates a text describing the image content. We present the overall pipeline in Figure~\ref{figure:pipeline} and explain the details of the pipeline in the following sections.

\subsection{Text Reconstruction Training}
\label{method:train}

\paragraph{\textbf{Region Noise Injection}} The text reconstruction task aims to train our framework to generate text based on the CLIP text features $T_c$, as illustrated in Figure \ref{figure:pipeline}. Our observation on the CLIP Embedding space demonstrates that the gap between the text embedding and subregion image representation satisfies a Gaussian distribution. To mitigate the gap and maintain a consistent format with visual features in inference, we propose region-aware noise injection.
In detail, We first encode $t$ from text corpus $T$ with CLIP text encoder to get text features $T_c$. The $T_c$ is repeated $N_{cr}$ times and added different noise $n_i \in R^{D}, i\in\{1... N_{cr}\}$ from a uniform distribution with zero means and $\sigma$ variance. We apply L2 normalization to the resulting text region feature $T_{cr} \in R^{N_{cr} \times D}$. The process is formulated as follows: 
\begin{align}
    &T_c = \mathrm{CLIP}(t) \in R^{D} \\
    &T_{cr} = \mathrm{Concat}(T_c, T_c, \dots T_c) \in R^{N_{cr} \times D} \\
    &T_{cr}^{i} = \mathrm{L2Norm}(T_{cr}^{i} + n_i) \quad n_i \sim \mathcal{N}(0,\,\sigma^{2})
\end{align}
where $\mathrm{L2Norm}$ is a $l2$-normalization and $\mathrm{Concat}$ is the concatenation operation. The elements in $T_{cr}$ form a cluster of points in the CLIP embedding space centered around $T_c$, which represent captions semantically similar to caption $t$. 

\paragraph{\textbf{Adaptor Decoder}}
The adaptor decoder is designed to project the feature in the CLIP embedding space to the language model embedding space, enabling the language model to generate text based on image or text features in CLIP. Specifically, we have a set of learnable queries $Q \in R^{N_q \times D}$, a transformer decoder \cite{Vaswani2017AttentionIA}, and an MLP module. The learnable queries $Q$ are first updated by self-attention and then fed into a cross-attention module with $T_{cr}$ as the input key and value. The output feature is processed by a feed-forward network to obtain updated learnable queries $Q' \in R^{N_q \times D}$. Finally, $Q'$ is projected by the MLP module to get the prefix embedding $E \in R^{N_q \times D_l}$. The process is formulated as follows:
\begin{align}
    &Q = \mathrm{SelfAttn}(Q) \in R^{N_{q} \times D}  \\
    &Q = \mathrm{CrossAttn}(Q, T_{cr})\in R^{N_{q} \times D} \\
    &Q' = \mathrm{FFN}(Q)\in R^{N_{q} \times D} \\
    &E = \mathrm{MLP}(Q')\in R^{N_{q} \times D_l}
\end{align}
Through cross-attention in the adaptor decoder, the learnable queries $Q$ adaptively select informative parts in $T_{cr}$. At last, the language model generates the input text $t$ based on the prefix embedding $E$. Our objective can be described as:
\begin{align}
   L_{recons}(\theta) = - \frac{1}{|t|} \sum_{i=1}^{|t|} {\log P_{\theta}(w_i|w_{<i}, E, \theta_{c}, \theta_{l}})
\end{align}
where $w_i$ is the $i^{th}$ words in $t$.

\subsection{Zero-shot Caption Generation}




\paragraph{\textbf{Sub-region Feature Aggregation}} 
\label{method:subregion}
Zero-shot caption generation aims to generate captions with a text-only trained adaptor decoder, CLIP, and a language model. Based on our observation that image subregion features exhibit higher similarity to caption features, we propose sub-region feature aggregation to integrate the image global information with subregion information.
Specifically, the ViT-based visual encoder processes images by dividing them into patches and incorporates a class token. In the last layer of ViT, we define the image patch features as $I_{p} \in R^{(N_p + 1) \times D_v}$, where $N_p$ is the number of image patches, $D_v$ is the dimensionality of ViT and the first element in $I_{p}$ is the class token $I_{p}[0]  \in R^{D}$. 
The global image feature $I_{c} \in R^{D}$ is obtained by a linear projection on class token $I_{p}[0]$. We select the patches with the top $N_{cr}$ score in the class token's attention weight and denote them as informative patch tokens. The subregion features $I_{s} \in R^{N_{cr} \times D} $ are obtained by aggregating patch features in informative patch based on corresponding attention weight $A \in R^{N_{cr} \times (N_p + 1)}$. Finally, the subregion-enhanced image feature $I_{cr} \in R^{N_{cr} \times D}$ is acquired by taking the average of $I_{c}$ and $I_{s}$. The process can be formulated as follows:
\begin{align}
    &I_{p}' = \mathrm{Linear}(I_{p}) \in R^{(N_p + 1) \times D}  \\
    &I_{c} =  I_{p}'[0] \in R^{D} \\
    &I_{s} = A  I_{p}' \in R^{N_{cr} \times D} \\
    &I_{cr} = \mathrm{Concat}(I_{s}^1 + I_{c}, \dots, I_{s}^{N_{cr}} + I_{c})  \in R^{N_{q} \times D}
\end{align}
where $I_{p}'$ is the projected image patch features, and $\mathrm{Linear}$ is the linear projection that projects visual features to CLIP multimodal embedding space. The $I_{cr}$ represents the image feature in CLIP space and is used to generate text in the same way as text region feature $T_{cr}$.


\begin{figure}[t!]
  \centering
  \includegraphics[width=0.45\textwidth]{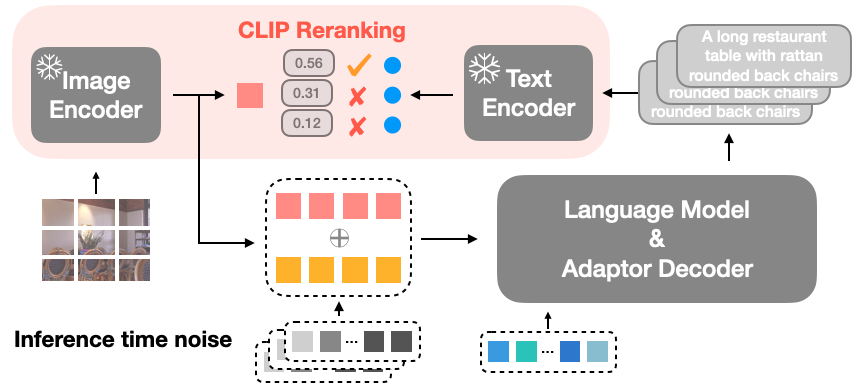}
   \caption{\textbf{Multiple sampling and filtering pipeline} During inference, each image uses noise to generate several different captions, which are reranked by CLIP to output the best.}
    \label{figure:rerank}
\end{figure}

\paragraph{\textbf{Multiple Sampling and Filtering}} 
The \textit{modality gap} is alleviated by the region noise injection in text reconstruction training, however, the noise introduces additional uncertainty. We propose a \textit{multiple sampling and filtering} strategy that incorporates inference-time noise and CLIP reranking to address the issue and boost performance, which is illustrated in Figure~\ref{figure:rerank}. Specifically, we introduce inference time noise sampled from a normal distribution $\mathcal{N}(0,\sigma^{2})$ into the subregion enhanced image feature $I_{cr} \in R^{N_{cr} \times D}$. We generate text based on the perturbed subregion enhanced image feature $I_{cr}$, and repeat this process $S$ times to generate $S$ diverse texts. CLIP is utilized to evaluate the cosine similarity between the generated texts and the image. We select the text with the highest similarity as the predicted image caption. This strategy leverages CLIP knowledge to improve the generation quality of our model.

\subsection{Zero-shot Visual Question Answering with Captioning}
In this part, we illustrate the potential extensibility by showing the pipeline for zero-shot VQA with text-only trained MacCap. As there is no supervision from VQA data, we convert the image into a caption by MacCap. To answer the question about the image, we only use the LLM in MacCap to do open-end text generation based on a VQA prompt. The prompt contains the question and image caption, which is \textit{"[caption] Question: [question] Answer:"}. Given the inherent difficulty of the zero-shot VQA task, we transform the VQA task into an image-text retrieval task. Specifically, the generated answer is embedded by the CLIP text encoder and computed cosine similarity with CLIP text embedding from answer candidates.
\begin{table*}
\tabcolsep=2pt
\small
\centering
\begin{tabular}{ccccccccccccccccccc}
\toprule
\toprule
\multicolumn{1}{c|}{\multirow{2}{*}{Method}} & \multicolumn{6}{c|}{CC3M to MS-COCO}                                                                   & \multicolumn{6}{c|}{MS-COCO to Flickr30k}                          & \multicolumn{6}{c}{Flickr30k to MS-COCO}      \\
\multicolumn{1}{c|}{}                        & B@1 & B@4                        & M     & R\_L                   & C     & \multicolumn{1}{c|}{S}     & B@1   & B@4   & M     & R\_L  & C     & \multicolumn{1}{c|}{S}     & B@1   & B@4   & M     & R\_L  & C     & S     \\ \hline
\multicolumn{1}{c|}{CLIPRe}                  & -     & 0.046 & 0.133 & -     & 0.256 & \multicolumn{1}{c|}{0.092} & 0.387 & 0.044 & 0.096 & 0.272 & 0.059 & \multicolumn{1}{c|}{0.042} & 0.414 & 0.052 & 0.125 & 0.307 & 0.183 & 0.057 \\
\multicolumn{1}{c|}{ZeroCap}                 & -     & 0.026 & 0.115 & -     & 0.146 & \multicolumn{1}{c|}{0.055} & -     & -     & -     & -     & -     & \multicolumn{1}{c|}{-}     & -     & -     & -     & -     & -     & -\\
\multicolumn{1}{c|}{MAGIC}                   & -     & -     & -     & -     & -     & \multicolumn{1}{c|}{-}     & 0.464 & 0.062 & 0.122 & 0.313 & 0.175 & \multicolumn{1}{c|}{-}     & 0.414 & 0.052 & 0.125 & 0.307 & 0.183 & -     \\
\multicolumn{1}{c|}{CapDec}                  & -     & -     & -     & -     & -     & \multicolumn{1}{c|}{-}     & 0.602 & 0.173 & 0.186 & 0.427 & 0.357 & \multicolumn{1}{c|}{-}     & 0.433 & 0.092 & 0.163 & 0.367 & 0.273 & -     \\
\multicolumn{1}{c|}{DeCap}                   & -     & 0.088 & 0.160 & -     & 0.421 & \multicolumn{1}{c|}{0.109} & -     & 0.163 & 0.179 & -     & 0.357 & \multicolumn{1}{c|}{0.111} & -     & 0.121 & 0.180 & -     & 0.444 & 0.109 \\ \hline
\multicolumn{19}{c}{Frozen Language Model}                                                                                                                        \\ \hline
\multicolumn{1}{c|}{Baseline}                  & 0.318 & 0.034 & 0.094 & 0.221 & 0.101 & \multicolumn{1}{c|}{0.046} & 0.429 & 0.072 & 0.126 & 0.305 & 0.140 & \multicolumn{1}{c|}{0.067} & 0.375 & 0.049 & 0.118 & 0.296 & 0.112 & 0.055 \\
\multicolumn{1}{c|}{$\mathrm{MAGIC}^\dagger$}  & 0.188 & 0.004 & 0.054 & 0.142 & 0.021 & \multicolumn{1}{c|}{0.011} & 0.188 & 0.006 & 0.051 & 0.134 & 0.021 & \multicolumn{1}{c|}{0.013} & 0.188 & 0.004 & 0.054 & 0.142 & 0.021 & 0.011 \\
\multicolumn{1}{c|}{$\mathrm{CapDec}^\dagger$} & 0.372 & 0.046 & 0.116 & 0.288 & 0.093 &\multicolumn{1}{c|}{0.052}  & 0.490 & 0.105 & 0.153 & 0.373 & 0.183 & \multicolumn{1}{c|}{0.090} & 0.453 & 0.087 & 0.151 & 0.353 & 0.178 & 0.064 \\
\multicolumn{1}{c|}{$\mathrm{DeCap}^\dagger$}  & 0.462 & 0.089 & 0.152 & 0.341 & 0.292 & \multicolumn{1}{c|}{0.093}& 0.511 & 0.099 & 0.153 & 0.362 & 0.247 & \multicolumn{1}{c|}{0.087} & 0.455 & 0.088 & 0.162 & 0.360 & 0.273 & \textbf{0.101} \\
\rowcolor{blue!6}\multicolumn{1}{c|}{MacCap}                  & \textbf{0.591} & \textbf{0.176} & \textbf{0.200} & \textbf{0.443} & \textbf{0.525} & \multicolumn{1}{c|}{\textbf{0.120}}  & \textbf{0.595} & \textbf{0.154} & \textbf{0.179} & \textbf{0.413} & \textbf{0.303} & \multicolumn{1}{c|}{\textbf{0.114}} & \textbf{0.473} & \textbf{0.092} & \textbf{0.166} & \textbf{0.362} & \textbf{0.278} & 0.092 \\ \toprule
\end{tabular}
\caption{\textbf{Zero-shot Cross Domain Captioning}: We conduct experiments on cross-domain image captioning tasks. X to Y means source to target domain. We reproduce Magic, CapDec, and DeCap under the frozen LLM setting and mark them with \textbf{$\dagger$}.}
\label{tab:cross_domain_results}
\end{table*}

\section{Experiments}

\subsection{Experimental setting}

\paragraph{\textbf{Datasets and Evaluation}} 
Our experimental evaluations are performed on two common benchmark datasets for image captioning: MSCOCO \cite{mscoco} and Flickr30k \cite{Flickr30k}. The MSCOCO dataset contains over 11,000 images, each associated with five captions. We follow previous works \cite{CapDec} using the widely-used Karpathy et al. split, which partitions the dataset into 5000 images for validation and 5,000 for testing. The Flickr30k dataset consists of 29,000 images for training and 1,000 images for testing. For training, we use the texts from MSCOCO, Flickr30K, and CC3M \cite{cc3m} datasets. Following previous works \cite{DeCap}, we remove the sentences with more than fifteen words. The resulting training text corpus has 566,747 sentences for MSCOCO, 144,998 sentences for Flickr30K, and 3,302,783 sentences for CC3M. To evaluate the performance of our models, we report the results on four standard evaluation metrics for captioning: BLEU \cite{Papineni2002BleuAM}, METEOR \cite{Banerjee2005METEORAA}, CIDEr \cite{Vedantam2014CIDErCI}, and SPICE \cite{Anderson2016SPICESP}. Additionally, we use the popular visual question answering benchmark VQAV2 \cite{vqav2} to evaluate the model's ability on complex visual tasks. The number of answer candidates for VQAV2 is 3,128. We randomly chose 20,000 samples from VQAV2 validation set for testing. 


\paragraph{\textbf{Implementation Details}} 
For a fair comparison with previous works \cite{CapDec,MAGIC,ZeroCap,DeCap}, we employ a frozen Vit-B/32 CLIP model. The adaptor decoder consists of one layer Transformer Decoder \cite{Vaswani2017AttentionIA} with 8 attention heads. For text reconstruction training, we set the noise variance $\sigma$ to 0.016 as suggested in \cite{CapDec}, and the region concept feature length $N_{cr}$ is set to 10. In caption generation, the sampling number $S$ in inference is set to 20. The text generation strategy is beam search with 4 beams. For the language model, we adopt a frozen pre-trained OPT \cite{Zhang2022OPTOP} 1.3b model. Our model and the reproduced baseline are trained with a batch size of 128 and a learning rate of 4e-4.

\paragraph{\textbf{Baselines}} 
The following zero-shot captioning methods are compared in this study. \textbf{ZeroCap} leverages CLIP and GPT-2 to solve an optimization problem iteratively during inference. \textbf{DeCap} \cite{DeCap} utilizes a memory bank to project image embeddings into the text embedding space. We sample 50M texts in CC3M \cite{cc3m} datasets to generate the memory bank for DeCap. Additionally, we define \textbf{Baseline} as the model trained in the same way as DeCap but using image embedding directly in inference.
\textbf{MAGIC} \cite{MAGIC} incorporates a CLIP-induced score during inference to influence the language model's caption generation process. We show the performance of MAGIC with a fine-tuned language model as reported in \cite{MAGIC} and the performance of MAGIC with a frozen pre-trained language model. \textbf{CLIPRe} is a retrieval-based baseline mentioned in \cite{MAGIC}. \textbf{CapDec} \cite{CapDec} applies noise injection strategy in text reconstruction training, enabling the direct use of visual embedding in inference. We use the MLP variant of CapDec model. \textbf{SM} \cite{Zeng2022SocraticMC} is a modular framework in which multiple pre-trained models may be composed zero-shot. It uses the GPT-3 \cite{gpt3} API from OpenAI and achieves favorable performance.

\begin{table*}
\tabcolsep=2pt
\small
\centering
\begin{threeparttable}
\begin{tabular}{cccccccccccccccc}
\toprule
\toprule
\multicolumn{1}{c|}{\multirow{2}{*}{Method}} & \multicolumn{3}{c|}{Data}                                 & \multicolumn{6}{c|}{MSCOCO}                                        & \multicolumn{6}{c}{Flickr30k}                 \\
\multicolumn{1}{c|}{}                        & P.         & I.         & \multicolumn{1}{c|}{T.}         & $B\texttt{@}1$   & $B\texttt{@}4$   & M     & $R_L$   & C     & \multicolumn{1}{c|}{S}     & $B\texttt{@}1$   & $B\texttt{@}4$   & M     & $R_L$   & C     & S     \\ \hline
\multicolumn{1}{c|}{CLIPCap \cite{clipcap}}                 & \checkmark &            & \multicolumn{1}{c|}{}           & -     & 0.335 & 0.275 & -     & 1.131 & \multicolumn{1}{c|}{0.211} & -     & -     & -     & -     & -     & -     \\
\multicolumn{1}{c|}{CLIP-VL \cite{Shen2021HowMC}}                 & \checkmark &            & \multicolumn{1}{c|}{}           & -     & 0.375 & 0.281 & -     & 1.231 & \multicolumn{1}{c|}{0.219} & -     & -     & -     & -     & -     & -     \\
\multicolumn{1}{c|}{UVC-VI \cite{Liu2021AligningSV}}                  & \checkmark &            & \multicolumn{1}{c|}{}           & -     & 0.220 & 0.214 & -    & 0.723 & \multicolumn{1}{c|}{}      & -     & -     & -     & -     & -     & -     \\
\multicolumn{1}{c|}{Barraco et al. \cite{Barraco2022TheUE}}          & \checkmark &            & \multicolumn{1}{c|}{}           & -     & 0.360 & 0.278 & -     & 1.149 & \multicolumn{1}{c|}{0.208} & -     & -     & -     & -     & -     & -     \\
\multicolumn{1}{c|}{ESPER-Style \cite{Yu2022MultimodalKA}}             &            & \checkmark & \multicolumn{1}{c|}{\checkmark} & -     & 0.219 & 0.219 & -     & 0.782 & \multicolumn{1}{c|}{}      & -     & -     & -     & -     & -     & -     \\
\multicolumn{1}{c|}{ESPER-Free \cite{Yu2022MultimodalKA}}              &            & \checkmark & \multicolumn{1}{c|}{}           & -     & 0.063 & 0.133 & -     & 0.291 & \multicolumn{1}{c|}{}      & -     & -     & -     & -     & -     & -     \\
\multicolumn{1}{c|}{ZeroCap$^{\star} $\cite{ZeroCap}}        &            &            & \multicolumn{1}{c|}{\checkmark} & 0.498 & 0.007 & 0.154 & 0.318 & 0.345 & \multicolumn{1}{c|}{0.092} & 0.447 & 0.054 & 0.118 & 0.273 & 0.168 & 0.062 \\
\multicolumn{1}{c|}{CLIPRe \cite{MAGIC}}                  &            &            & \multicolumn{1}{c|}{\checkmark} & 0.395 & 0.049 & 0.114 & 0.290 & 0.136 & \multicolumn{1}{c|}{0.053} & 0.385 & 0.052 & 0.116 & 0.276 & 0.100 & 0.057 \\
\multicolumn{1}{c|}{MAGIC \cite{MAGIC}}                   &            &            & \multicolumn{1}{c|}{\checkmark} & 0.568 & 0.129 & 0.174 & 0.399 & 0.493 & \multicolumn{1}{c|}{0.113} & 0.445 & 0.064 & 0.131 & 0.316 & 0.204 & 0.071 \\
\multicolumn{1}{c|}{CapDec \cite{CapDec}}                   &            &            & \multicolumn{1}{c|}{\checkmark} & 0.692& 0.264 & 0.251 & 0.518 & 0.918 & \multicolumn{1}{c|}{-} & 0.555     & 0.177 & 0.200 & 0.439 & 0.391 & - \\
\multicolumn{1}{c|}{DeCap \cite{DeCap}}                   &            &            & \multicolumn{1}{c|}{\checkmark} & -     & 0.247 & 0.250 & -     & 0.912 & \multicolumn{1}{c|}{0.187} & -     & 0.212 & 0.218 & -     & 0.567 & 0.152 \\
\hline
\multicolumn{16}{c}{Frozen Language Model}                                                                                                                                                                                    \\ \hline
\multicolumn{1}{c|}{Baseline}                &            &            & \multicolumn{1}{c|}{\checkmark} & 0.414 & 0.069 & 0.141 & 0.317 & 0.221 & \multicolumn{1}{c|}{0.079} & 0.418 & 0.069 & 0.127 & 0.308 & 0.136 & 0.070 \\
\multicolumn{1}{c|}{$\mathrm{MAGIC}^\dagger $\cite{MAGIC}}                   &            &            & \multicolumn{1}{c|}{\checkmark} & 0.188 & 0.004 & 0.054 & 0.042 & 0.021 & \multicolumn{1}{c|}{0.011} & 0.188 & 0.006 & 0.051 & 0.134 & 0.021 & 0.013 \\
\multicolumn{1}{c|}{$\mathrm{CapDec}^\dagger $\cite{CapDec}}                  &            &            & \multicolumn{1}{c|}{\checkmark} & 0.537 & 0.156 & 0.206 & 0.435 & 0.465 & \multicolumn{1}{c|}{0.134} & 0.429 & 0.072 & 0.136 & 0.336 & 0.127 & 0.054 \\
\multicolumn{1}{c|}{$\mathrm{DeCap}^\dagger $\cite{DeCap}}                   &            &            & \multicolumn{1}{c|}{\checkmark} & 0.531 & 0.125 & 0.188 & 0.403 & 0.427 & \multicolumn{1}{c|}{0.126} & 0.485 & 0.096 & 0.143 & 0.351 & 0.213 & 0.079 \\
\rowcolor{blue!6}\multicolumn{1}{c|}{MacCap}                    &            &            & \multicolumn{1}{c|}{\checkmark} & \textbf{0.614} & \textbf{0.174} & \textbf{0.223} & \textbf{0.459} & \textbf{0.697} & \multicolumn{1}{c|}{\textbf{0.157}} & \textbf{0.564} & \textbf{0.153} & \textbf{0.189} & \textbf{0.414} & \textbf{0.358} & \textbf{0.124} \\ \toprule
\end{tabular}
\end{threeparttable}
\caption{\textbf{Zero-shot In Domain Captioning}: The notation "P.", "I.", and "T." are used to represent paired data, unpaired image data, and unpaired text data, respectively. We reproduce Magic, CapDec, and DeCap under frozen language model setting and mark them with \textbf{$\dagger$}. Results tagged  $\star$ are from \cite{MAGIC}}
\label{tab:in_domain_results}
\end{table*}

\begin{table}{}{}
\small
\centering
\tabcolsep=4pt
\begin{tabular}{clllllll}
\toprule
\toprule
\multicolumn{1}{c|}{\multirow{2}{*}{Method}}            &  & \multicolumn{3}{c|}{VQAV2 Val (\%)} & \multicolumn{3}{c}{MSCOCO}                    \\
\multicolumn{1}{c|}{}                                   &              & Top1  & Top5   & \multicolumn{1}{c|}{Top10} & $R_L$  & C     & S     \\ \hline
\multicolumn{8}{c}{\textit{Finetuned Language Model}}                                                                                                      \\ \hline
\multicolumn{1}{c|}{{$\mathrm{CapDec}^\dagger$}}        &              & 0.86 & 3.13 & \multicolumn{1}{c|}{3.71} & \textbf{0.518} & \textbf{0.918} & - \\
\multicolumn{1}{c|}{{$\mathrm{DeCap}^\dagger$}}         &              & 4.09 & 10.89 & \multicolumn{1}{c|}{14.33} & - & 0.912 & \textbf{0.187}  \\ \hline
\multicolumn{8}{c}{\textit{Frozen Language Model}}                                                                                                     \\ \hline
\multicolumn{1}{c|}{Baseline}                           &              & 3.26 & 7.24 & \multicolumn{1}{c|}{11.21} & - & - & - \\
\multicolumn{1}{c|}{CapDec}                             &              & 6.53 & 11.06 & \multicolumn{1}{c|}{15.00} & 0.435 & 0.465 & 0.134 \\
\multicolumn{1}{c|}{DeCap}                              &              & 6.00 & 11.81 & \multicolumn{1}{c|}{15.57} & 0.403 & 0.427 & 0.126 \\
\rowcolor{blue!6}\multicolumn{1}{c|}{MacCap}                             &              & \textbf{7.96} & \textbf{14.00} & \multicolumn{1}{c|}{\textbf{18.72}} & 0.459 & 0.697 & 0.157 \\
\toprule
\end{tabular}
\caption{\textbf{Zero-shot VQA results} on VQAV2 validation set. $\dagger$ means the models come from the official release. Our method achieves superior performance under a frozen language model setting.}
\label{table:VQAV2}
\end{table}

\subsection{Zero-Shot Cross Domain Captioning}

In this section, we present a comprehensive evaluation of our method in the zero-shot cross-domain captioning setting. The zero-shot means our model is trained with only text data and the cross-domain means the training caption texts are different from captions in the test dataset. The CC3M is web-scale noisy captions data while MSCOCO and Flickr30K are human-annotated high-quality caption data. 
We compare our approach with previous methods on three different cross-domain settings to assess its performance and generalizability. We show our cross-domain image captioning results in Table~\ref{tab:cross_domain_results}. We have observed a positive correlation between the size of the training text corpus and the performance gain of MacCap. Our method achieves superiority in both domains in most metrics under our frozen language model setting. 



\subsection{Zero-Shot In Domain Image Captioning}
\paragraph{\textbf{Setting}} In this section, we conduct zero-shot in-domain image captioning experiments, where the models are trained and tested on the same domain. We compare our method with other supervised methods, unpaired image captioning methods, and text-only training methods. 

\paragraph{\textbf{Results}} We show our results on MSCOCO \cite{mscoco} and Flickr30K \cite{Flickr30k} in Table~\ref{tab:in_domain_results}. Our method outperforms other methods on both domains in all metrics under our frozen language model setting and shows a gain of +27 in CIDEr compared with DeCap \cite{DeCap}. We also achieve higher performance compared with fully supervised methods such as Laina et al. \cite{Laina2019TowardsUI} and Feng et al \cite{Feng2018UnsupervisedIC} in terms of CIDEr on MSCOCO.

\subsection{Zero-Shot Visual Question Answering}
In this section, we conduct zero-shot visual question-answering experiments. Due to the redundancy of the large language model generation results, we do not use the VQAV2 traditional evaluation metric where the predicted answer should be the same as the ground truth answer but instead use an image-text retrieval task where the answer candidates are provided. We report the top 1,5 and 10 accuracies in $\%$ for 20,000 VQAV2 validation set samples.
Compared with previous zero-shot captioning methods \cite{DeCap,CapDec}, including a baseline where the language model generates answers solely based on the question without a caption, our results are presented in Table~\ref{table:VQAV2}. Our method outperforms other methods with and without a frozen language model. We observe a performance degradation between the frozen language model and the finetuned language model, which indicates that the finetune language model on zero-shot captioning model harms model performance on other tasks.

\subsection{Ablation Study}
In this section, we conduct an ablation study to provide a comprehensive interpretation of our proposed methods. The experiments are conducted under zero-shot cross-domain image captioning settings where the model is trained on CC3M text corpus and evaluated on Flickr30K dataset.

\paragraph{\textbf{Text Reconstruction Training}}
To validate the effectiveness of the region noise design in our framework, we conducted experiments to determine whether the observed improvements were due to the noise injection or the sequential representation of the text. we modify the input text feature of the adaptor decoder module, where we define two modes: \textit{single token} and \textit{multiple token}. In the \textit{single token} mode, a single text embedding $T_c$ is provided as input to the adaptor decoder. In contrast, in the \textit{Multiple tokens} mode, multiple text embeddings are used as input for the adaptor decoder.
The \textit{w/ noise} or \textit{w/o noise} indicate whether adding noise to the embedding. We present the results in Table~\ref{table:ablation}. Based on the results, we can conclude that the strong performance of our approach can be attributed to the effective combination of noise injection and sequential representation.
\begin{table}{}{}
\small
\centering
\tabcolsep=2pt
\begin{tabular}{clllllll}
\toprule
\toprule
\multicolumn{1}{c|}{\multirow{2}{*}{Method}} &  & \multicolumn{6}{c}{Flickr30K}                    \\
\multicolumn{1}{c|}{}                        &                           & $B\texttt{@}1$   & $B\texttt{@}4$   & M     & $R_L$   & C     & S     \\ \hline
\multicolumn{8}{c}{\textit{Text Reconstruction Training}}                                                                                                      \\ \hline
\multicolumn{1}{c|}{\textit{single token w/o noise}}                &              &    0.396    & 0.055 & 0.123 &   0.277    & 0.147 & 0.070 \\
\multicolumn{1}{c|}{\textit{single token w/ noise}}                 &              &    0.445   &  0.082 & 0.110 &  0.282     & 0.143 & 0.065 \\ 
\multicolumn{1}{c|}{\textit{multiple token w/o noise}}              &              &     0.388   & 0.049 & 0.122 &  0.277     & 0.135 &  0.060 \\
\multicolumn{1}{c|}{\textit{multiple token w/ noise}}                 &              &     0.520 & 0.107 &   0.153    & 0.369 & 0.197 & 0.084  \\ \hline
\multicolumn{8}{c}{\textit{Zero-shot Caption Generation}}                                                                                                     \\ \hline
\multicolumn{1}{c|}{\textit{CLS token}}                   &                 & 0.493 & 0.096 & 0.134 & 0.350 & 0.140 & 0.064 \\
\multicolumn{1}{c|}{\textit{subregion aggregation}}                  &                 & 0.520 & 0.107 & 0.153 & 0.369 & 0.197 & 0.084 \\
\multicolumn{1}{c|}{\textit{sampling and filtering}}                   &                 & \textbf{0.542} & \textbf{0.130} & \textbf{0.152} & \textbf{0.378} &  \textbf{0.199} & \textbf{0.078} \\
\toprule
\end{tabular}
\caption{\textbf{Ablation Results} on Flickr30K datasets. We evaluate the effectiveness of our training and inference paradigms.}
\label{table:ablation}
\end{table}

\paragraph{\textbf{Zero-shot Caption Generation}}
We conduct an ablation study based on the model trained with region noise. We modified the input image feature of the adaptor decoder module. We defined two modes: \textit{CLS token} indicate $I_{cr}$ doesn't contain subregion features $I_s$ and \textit{subregion aggregation} indicate $I_{cr}$ is the sum of $I_s$ and $I_c$. Furthermore, we incorporated the \textit{sampling and filtering} strategy. The obtained outcomes are illustrated in Table~\ref{table:ablation}. Our observations reveal that the adoption of the \textit{sampling and filtering} approach led to a noteworthy improvement in the BLEU metric, signifying the rectification of erroneous instances within the generated captions. However, its impact on semantic comprehension and contextual coherence was relatively modest in comparison.


\section{Conclusion}

We present an in-depth analysis of the \textit{modality gap} phenomenon in the CLIP latent space, uncovering two key phenomena: tighter proximity of CLIP visual features within image subregions to paired captions and a modality gap adhering to a zero-mean Gaussian distribution. In response to these insights, we introduced a novel approach—region noise-injected text reconstruction training. Leveraging subregion feature aggregation in zero-shot caption generation, we harnessed subregion information to create a closer visual representation with paired caption representation. Additionally, we propose an inference-time noise and CLIP reranking to further boost performance. Experimental results demonstrate that MacCap outperforms state-of-the-art methods.

\bibliography{aaai24}

\begin{thebibliography}{55}
\providecommand{\natexlab}[1]{#1}

\bibitem[{Agrawal et~al.(2015)Agrawal, Lu, Antol, Mitchell, Zitnick, Parikh,
  and Batra}]{Agrawal2015VQAVQ}
Agrawal, A.; Lu, J.; Antol, S.; Mitchell, M.; Zitnick, C.~L.; Parikh, D.; and
  Batra, D. 2015.
\newblock VQA: Visual Question Answering.
\newblock \emph{International Journal of Computer Vision}, 123: 4--31.

\bibitem[{Alayrac et~al.(2022)Alayrac, Donahue, Luc, Miech, Barr, Hasson, Lenc,
  Mensch, Millican, Reynolds et~al.}]{flamingo}
Alayrac, J.-B.; Donahue, J.; Luc, P.; Miech, A.; Barr, I.; Hasson, Y.; Lenc,
  K.; Mensch, A.; Millican, K.; Reynolds, M.; et~al. 2022.
\newblock Flamingo: a visual language model for few-shot learning.
\newblock \emph{Advances in Neural Information Processing Systems}, 35:
  23716--23736.

\bibitem[{Anderson et~al.(2016)Anderson, Fernando, Johnson, and
  Gould}]{Anderson2016SPICESP}
Anderson, P.; Fernando, B.; Johnson, M.; and Gould, S. 2016.
\newblock SPICE: Semantic Propositional Image Caption Evaluation.
\newblock In \emph{European Conference on Computer Vision}.

\bibitem[{Anderson et~al.(2017)Anderson, He, Buehler, Teney, Johnson, Gould,
  and Zhang}]{conimgcap1}
Anderson, P.; He, X.; Buehler, C.; Teney, D.; Johnson, M.; Gould, S.; and
  Zhang, L. 2017.
\newblock Bottom-Up and Top-Down Attention for Image Captioning and Visual
  Question Answering.
\newblock \emph{2018 IEEE/CVF Conference on Computer Vision and Pattern
  Recognition}, 6077--6086.

\bibitem[{Antol et~al.(2015)Antol, Agrawal, Lu, Mitchell, Batra, Zitnick, and
  Parikh}]{vqav2}
Antol, S.; Agrawal, A.; Lu, J.; Mitchell, M.; Batra, D.; Zitnick, C.~L.; and
  Parikh, D. 2015.
\newblock Vqa: Visual question answering.
\newblock In \emph{Proceedings of the IEEE international conference on computer
  vision}, 2425--2433.

\bibitem[{Banerjee and Lavie(2005)}]{Banerjee2005METEORAA}
Banerjee, S.; and Lavie, A. 2005.
\newblock METEOR: An Automatic Metric for MT Evaluation with Improved
  Correlation with Human Judgments.
\newblock In \emph{IEEvaluation@ACL}.

\bibitem[{Barraco et~al.(2022)Barraco, Cornia, Cascianelli, Baraldi, and
  Cucchiara}]{Barraco2022TheUE}
Barraco, M.; Cornia, M.; Cascianelli, S.; Baraldi, L.; and Cucchiara, R. 2022.
\newblock The Unreasonable Effectiveness of CLIP Features for Image Captioning:
  An Experimental Analysis.
\newblock \emph{2022 IEEE/CVF Conference on Computer Vision and Pattern
  Recognition Workshops (CVPRW)}, 4661--4669.

\bibitem[{Brown et~al.(2020)Brown, Mann, Ryder, Subbiah, Kaplan, Dhariwal,
  Neelakantan, Shyam, Sastry, Askell, Agarwal, Herbert-Voss, Krueger, Henighan,
  Child, Ramesh, Ziegler, Wu, Winter, Hesse, Chen, Sigler, Litwin, Gray, Chess,
  Clark, Berner, McCandlish, Radford, Sutskever, and Amodei}]{gpt3}
Brown, T.~B.; Mann, B.; Ryder, N.; Subbiah, M.; Kaplan, J.; Dhariwal, P.;
  Neelakantan, A.; Shyam, P.; Sastry, G.; Askell, A.; Agarwal, S.;
  Herbert-Voss, A.; Krueger, G.; Henighan, T.~J.; Child, R.; Ramesh, A.;
  Ziegler, D.~M.; Wu, J.; Winter, C.; Hesse, C.; Chen, M.; Sigler, E.; Litwin,
  M.; Gray, S.; Chess, B.; Clark, J.; Berner, C.; McCandlish, S.; Radford, A.;
  Sutskever, I.; and Amodei, D. 2020.
\newblock Language Models are Few-Shot Learners.
\newblock \emph{ArXiv}, abs/2005.14165.

\bibitem[{Changpinyo et~al.(2022)Changpinyo, Kukliansky, Szpektor, Chen, Ding,
  and Soricut}]{Changpinyo2022AllYM}
Changpinyo, S.; Kukliansky, D.; Szpektor, I.; Chen, X.; Ding, N.; and Soricut,
  R. 2022.
\newblock All You May Need for VQA are Image Captions.
\newblock In \emph{North American Chapter of the Association for Computational
  Linguistics}.

\bibitem[{Changpinyo et~al.(2021{\natexlab{a}})Changpinyo, Sharma, Ding, and
  Soricut}]{Changpinyo}
Changpinyo, S.; Sharma, P.; Ding, N.; and Soricut, R. 2021{\natexlab{a}}.
\newblock Conceptual 12M: Pushing Web-Scale Image-Text Pre-Training To
  Recognize Long-Tail Visual Concepts.
\newblock In \emph{2021 IEEE/CVF Conference on Computer Vision and Pattern
  Recognition (CVPR)}.

\bibitem[{Changpinyo et~al.(2021{\natexlab{b}})Changpinyo, Sharma, Ding, and
  Soricut}]{cc3m}
Changpinyo, S.; Sharma, P.~K.; Ding, N.; and Soricut, R. 2021{\natexlab{b}}.
\newblock Conceptual 12M: Pushing Web-Scale Image-Text Pre-Training To
  Recognize Long-Tail Visual Concepts.
\newblock \emph{2021 IEEE/CVF Conference on Computer Vision and Pattern
  Recognition (CVPR)}, 3557--3567.

\bibitem[{Das et~al.(2016)Das, Kottur, Gupta, Singh, Yadav, Moura, Parikh, and
  Batra}]{Das2016VisualD}
Das, A.; Kottur, S.; Gupta, K.; Singh, A.; Yadav, D.; Moura, J. M.~F.; Parikh,
  D.; and Batra, D. 2016.
\newblock Visual Dialog.
\newblock \emph{2017 IEEE Conference on Computer Vision and Pattern Recognition
  (CVPR)}, 1080--1089.

\bibitem[{Feng et~al.(2018)Feng, Ma, Liu, and Luo}]{Feng2018UnsupervisedIC}
Feng, Y.; Ma, L.; Liu, W.; and Luo, J. 2018.
\newblock Unsupervised Image Captioning.
\newblock \emph{2019 IEEE/CVF Conference on Computer Vision and Pattern
  Recognition (CVPR)}, 4120--4129.

\bibitem[{Guo et~al.(2022{\natexlab{a}})Guo, Li, Li, Tiong, Li, Tao, and
  Hoi}]{Guo2022FromIT}
Guo, J.; Li, J.; Li, D.; Tiong, A. M.~H.; Li, B.; Tao, D.; and Hoi, S.
  2022{\natexlab{a}}.
\newblock From Images to Textual Prompts: Zero-shot VQA with Frozen Large
  Language Models.
\newblock \emph{ArXiv}, abs/2212.10846.

\bibitem[{Guo et~al.(2022{\natexlab{b}})Guo, Zhang, Qiu, Ma, Miao, He, and
  Cui}]{calip}
Guo, Z.; Zhang, R.; Qiu, L.; Ma, X.; Miao, X.; He, X.; and Cui, B.
  2022{\natexlab{b}}.
\newblock CALIP: Zero-Shot Enhancement of CLIP with Parameter-free Attention.
\newblock \emph{ArXiv}, abs/2209.14169.

\bibitem[{Hudson and Manning(2019)}]{gqa}
Hudson, D.~A.; and Manning, C.~D. 2019.
\newblock GQA: A New Dataset for Real-World Visual Reasoning and Compositional
  Question Answering.
\newblock \emph{2019 IEEE/CVF Conference on Computer Vision and Pattern
  Recognition (CVPR)}, 6693--6702.

\bibitem[{Jia et~al.(2021)Jia, Yang, Xia, Chen, Parekh, Pham, Le, Sung, Li, and
  Duerig}]{align}
Jia, C.; Yang, Y.; Xia, Y.; Chen, Y.-T.; Parekh, Z.; Pham, H.; Le, Q.; Sung,
  Y.-H.; Li, Z.; and Duerig, T. 2021.
\newblock Scaling up visual and vision-language representation learning with
  noisy text supervision.
\newblock In \emph{International Conference on Machine Learning}, 4904--4916.
  PMLR.

\bibitem[{Laina, Rupprecht, and Navab(2019)}]{Laina2019TowardsUI}
Laina, I.; Rupprecht, C.; and Navab, N. 2019.
\newblock Towards Unsupervised Image Captioning With Shared Multimodal
  Embeddings.
\newblock \emph{2019 IEEE/CVF International Conference on Computer Vision
  (ICCV)}, 7413--7423.

\bibitem[{Li et~al.(2022)Li, Li, Xiong, and Hoi}]{blip}
Li, J.; Li, D.; Xiong, C.; and Hoi, S. C.~H. 2022.
\newblock BLIP: Bootstrapping Language-Image Pre-training for Unified
  Vision-Language Understanding and Generation.
\newblock In \emph{International Conference on Machine Learning}.

\bibitem[{Li et~al.(2021)Li, Zhang, Zhang, Yang, Li, Zhong, Wang, Yuan, Zhang,
  Hwang, Chang, and Gao}]{glip}
Li, L.~H.; Zhang, P.; Zhang, H.; Yang, J.; Li, C.; Zhong, Y.; Wang, L.; Yuan,
  L.; Zhang, L.; Hwang, J.-N.; Chang, K.-W.; and Gao, J. 2021.
\newblock Grounded Language-Image Pre-training.
\newblock \emph{2022 IEEE/CVF Conference on Computer Vision and Pattern
  Recognition (CVPR)}, 10955--10965.

\bibitem[{Li et~al.(2023)Li, Zhu, Wen, and Yang}]{DeCap}
Li, W.; Zhu, L.; Wen, L.; and Yang, Y. 2023.
\newblock DeCap: Decoding CLIP Latents for Zero-Shot Captioning via Text-Only
  Training.
\newblock \emph{arXiv preprint arXiv:2303.03032}.

\bibitem[{Liang et~al.(2022)Liang, Zhang, Kwon, Yeung, and Zou}]{MindGap}
Liang, V.~W.; Zhang, Y.; Kwon, Y.; Yeung, S.; and Zou, J.~Y. 2022.
\newblock Mind the gap: Understanding the modality gap in multi-modal
  contrastive representation learning.
\newblock \emph{Advances in Neural Information Processing Systems}, 35:
  17612--17625.

\bibitem[{Liu et~al.(2021)Liu, Wu, You, Ge, Zou, and Sun}]{Liu2021AligningSV}
Liu, F.; Wu, X.; You, C.; Ge, S.; Zou, Y.; and Sun, X. 2021.
\newblock Aligning Source Visual and Target Language Domains for Unpaired Video
  Captioning.
\newblock \emph{IEEE Transactions on Pattern Analysis and Machine
  Intelligence}, 44: 9255--9268.

\bibitem[{Liu et~al.(2023)Liu, Li, Wu, and Lee}]{llava}
Liu, H.; Li, C.; Wu, Q.; and Lee, Y.~J. 2023.
\newblock Visual Instruction Tuning.
\newblock \emph{ArXiv}, abs/2304.08485.

\bibitem[{Marino et~al.(2019)Marino, Rastegari, Farhadi, and Mottaghi}]{okvqa}
Marino, K.; Rastegari, M.; Farhadi, A.; and Mottaghi, R. 2019.
\newblock OK-VQA: A Visual Question Answering Benchmark Requiring External
  Knowledge.
\newblock \emph{2019 IEEE/CVF Conference on Computer Vision and Pattern
  Recognition (CVPR)}, 3190--3199.

\bibitem[{Mokady(2021)}]{clipcap}
Mokady, R. 2021.
\newblock ClipCap: CLIP Prefix for Image Captioning.
\newblock \emph{ArXiv}, abs/2111.09734.

\bibitem[{Ning et~al.(2023)Ning, Qiu, Liu, and He}]{hoiclip}
Ning, S.; Qiu, L.; Liu, Y.; and He, X. 2023.
\newblock HOICLIP: Efficient Knowledge Transfer for HOI Detection with
  Vision-Language Models.
\newblock \emph{ArXiv}, abs/2303.15786.

\bibitem[{Niu et~al.(2018)Niu, Zhang, Zhang, Zhang, Lu, and
  Wen}]{Niu2018RecursiveVA}
Niu, Y.; Zhang, H.; Zhang, M.; Zhang, J.; Lu, Z.; and Wen, J.-R. 2018.
\newblock Recursive Visual Attention in Visual Dialog.
\newblock \emph{2019 IEEE/CVF Conference on Computer Vision and Pattern
  Recognition (CVPR)}, 6672--6681.

\bibitem[{Nukrai, Mokady, and Globerson(2022)}]{CapDec}
Nukrai, D.; Mokady, R.; and Globerson, A. 2022.
\newblock Text-Only Training for Image Captioning using Noise-Injected CLIP.
\newblock \emph{arXiv preprint arXiv:2211.00575}.

\bibitem[{Papineni et~al.(2002)Papineni, Roukos, Ward, and
  Zhu}]{Papineni2002BleuAM}
Papineni, K.; Roukos, S.; Ward, T.; and Zhu, W.-J. 2002.
\newblock Bleu: a Method for Automatic Evaluation of Machine Translation.
\newblock In \emph{Annual Meeting of the Association for Computational
  Linguistics}.

\bibitem[{Patashnik et~al.(2021)Patashnik, Wu, Shechtman, Cohen-Or, and
  Lischinski}]{styleclip}
Patashnik, O.; Wu, Z.; Shechtman, E.; Cohen-Or, D.; and Lischinski, D. 2021.
\newblock StyleCLIP: Text-Driven Manipulation of StyleGAN Imagery.
\newblock \emph{2021 IEEE/CVF International Conference on Computer Vision
  (ICCV)}, 2065--2074.

\bibitem[{Plummer et~al.(2015)Plummer, Wang, Cervantes, Caicedo, Hockenmaier,
  and Lazebnik}]{Flickr30k}
Plummer, B.~A.; Wang, L.; Cervantes, C.~M.; Caicedo, J.~C.; Hockenmaier, J.;
  and Lazebnik, S. 2015.
\newblock Flickr30k Entities: Collecting Region-to-Phrase Correspondences for
  Richer Image-to-Sentence Models.
\newblock \emph{International Journal of Computer Vision}, 123: 74--93.

\bibitem[{Radford et~al.(2021)Radford, Kim, Hallacy, Ramesh, Goh, Agarwal,
  Sastry, Askell, Mishkin, Clark et~al.}]{clip}
Radford, A.; Kim, J.~W.; Hallacy, C.; Ramesh, A.; Goh, G.; Agarwal, S.; Sastry,
  G.; Askell, A.; Mishkin, P.; Clark, J.; et~al. 2021.
\newblock Learning transferable visual models from natural language
  supervision.
\newblock In \emph{International conference on machine learning}, 8748--8763.
  PMLR.

\bibitem[{Rao et~al.(2021)Rao, Zhao, Chen, Tang, Zhu, Huang, Zhou, and
  Lu}]{denclip}
Rao, Y.; Zhao, W.; Chen, G.; Tang, Y.; Zhu, Z.; Huang, G.; Zhou, J.; and Lu, J.
  2021.
\newblock DenseCLIP: Language-Guided Dense Prediction with Context-Aware
  Prompting.
\newblock \emph{2022 IEEE/CVF Conference on Computer Vision and Pattern
  Recognition (CVPR)}, 18061--18070.

\bibitem[{Shen et~al.(2021)Shen, Li, Tan, Bansal, Rohrbach, Chang, Yao, and
  Keutzer}]{Shen2021HowMC}
Shen, S.; Li, L.~H.; Tan, H.; Bansal, M.; Rohrbach, A.; Chang, K.-W.; Yao, Z.;
  and Keutzer, K. 2021.
\newblock How Much Can CLIP Benefit Vision-and-Language Tasks?
\newblock \emph{ArXiv}, abs/2107.06383.

\bibitem[{Su et~al.(2022)Su, Lan, Liu, Liu, Yogatama, Wang, Kong, and
  Collier}]{MAGIC}
Su, Y.; Lan, T.; Liu, Y.; Liu, F.; Yogatama, D.; Wang, Y.; Kong, L.; and
  Collier, N. 2022.
\newblock Language models can see: plugging visual controls in text generation.
\newblock \emph{arXiv preprint arXiv:2205.02655}.

\bibitem[{Tewel et~al.(2022)Tewel, Shalev, Schwartz, and Wolf}]{ZeroCap}
Tewel, Y.; Shalev, Y.; Schwartz, I.; and Wolf, L. 2022.
\newblock ZeroCap: Zero-Shot Image-to-Text Generation for Visual-Semantic
  Arithmetic.
\newblock In \emph{2022 IEEE/CVF Conference on Computer Vision and Pattern
  Recognition (CVPR)}.

\bibitem[{Tiong et~al.(2022)Tiong, Li, Li, Savarese, and
  Hoi}]{Tiong2022PlugandPlayVZ}
Tiong, A. M.~H.; Li, J.; Li, B.; Savarese, S.; and Hoi, S. C.~H. 2022.
\newblock Plug-and-Play VQA: Zero-shot VQA by Conjoining Large Pretrained
  Models with Zero Training.
\newblock \emph{ArXiv}, abs/2210.08773.

\bibitem[{Vaswani et~al.(2017)Vaswani, Shazeer, Parmar, Uszkoreit, Jones,
  Gomez, Kaiser, and Polosukhin}]{Vaswani2017AttentionIA}
Vaswani, A.; Shazeer, N.~M.; Parmar, N.; Uszkoreit, J.; Jones, L.; Gomez,
  A.~N.; Kaiser, L.; and Polosukhin, I. 2017.
\newblock Attention is All you Need.
\newblock In \emph{NIPS}.

\bibitem[{Vedantam, Zitnick, and Parikh(2014)}]{Vedantam2014CIDErCI}
Vedantam, R.; Zitnick, C.~L.; and Parikh, D. 2014.
\newblock CIDEr: Consensus-based image description evaluation.
\newblock \emph{2015 IEEE Conference on Computer Vision and Pattern Recognition
  (CVPR)}, 4566--4575.

\bibitem[{Vinyals et~al.(2016)Vinyals, Toshev, Bengio, and Erhan}]{mscoco}
Vinyals, O.; Toshev, A.; Bengio, S.; and Erhan, D. 2016.
\newblock Show and Tell: Lessons Learned from the 2015 MSCOCO Image Captioning
  Challenge.
\newblock \emph{IEEE Transactions on Pattern Analysis and Machine
  Intelligence}, 39: 652--663.

\bibitem[{Wang and Liu(2020)}]{Wang2020UnderstandingTB}
Wang, F.; and Liu, H. 2020.
\newblock Understanding the Behaviour of Contrastive Loss.
\newblock \emph{2021 IEEE/CVF Conference on Computer Vision and Pattern
  Recognition (CVPR)}, 2495--2504.

\bibitem[{Wang and Isola(2020)}]{Wang2020UnderstandingCR}
Wang, T.; and Isola, P. 2020.
\newblock Understanding Contrastive Representation Learning through Alignment
  and Uniformity on the Hypersphere.
\newblock \emph{ArXiv}, abs/2005.10242.

\bibitem[{Wang et~al.(2021)Wang, Yu, Yu, Dai, Tsvetkov, and
  Cao}]{Wang_Yu_Yu_Dai_Tsvetkov_Cao_2021}
Wang, Z.; Yu, J.; Yu, A.; Dai, Z.; Tsvetkov, Y.; and Cao, Y. 2021.
\newblock SimVLM: Simple Visual Language Model Pretraining with Weak
  Supervision.

\bibitem[{Xu et~al.(2015)Xu, Ba, Kiros, Cho, Courville, Salakhutdinov, Zemel,
  and Bengio}]{conimgcap4}
Xu, K.; Ba, J.; Kiros, R.; Cho, K.; Courville, A.~C.; Salakhutdinov, R.; Zemel,
  R.~S.; and Bengio, Y. 2015.
\newblock Show, Attend and Tell: Neural Image Caption Generation with Visual
  Attention.
\newblock In \emph{International Conference on Machine Learning}.

\bibitem[{Yao et~al.(2018)Yao, Pan, Li, and Mei}]{conimgcap3}
Yao, T.; Pan, Y.; Li, Y.; and Mei, T. 2018.
\newblock Exploring Visual Relationship for Image Captioning.
\newblock In \emph{European Conference on Computer Vision}.

\bibitem[{Yu et~al.(2022)Yu, Chung, Yun, Hessel, Park, Lu, Ammanabrolu,
  Zellers, Bras, Kim, and Choi}]{Yu2022MultimodalKA}
Yu, Y.; Chung, J.; Yun, H.; Hessel, J.; Park, J.~S.; Lu, X.; Ammanabrolu, P.;
  Zellers, R.; Bras, R.~L.; Kim, G.; and Choi, Y. 2022.
\newblock Multimodal Knowledge Alignment with Reinforcement Learning.
\newblock \emph{ArXiv}, abs/2205.12630.

\bibitem[{Yuksekgonul et~al.(2022)Yuksekgonul, Bianchi, Kalluri, Jurafsky, and
  Zou}]{clipbagofwords}
Yuksekgonul, M.; Bianchi, F.; Kalluri, P.; Jurafsky, D.; and Zou, J.~Y. 2022.
\newblock When and why vision-language models behave like bags-of-words, and
  what to do about it?
\newblock \emph{ArXiv}, abs/2210.01936.

\bibitem[{Zeng et~al.(2022)Zeng, Wong, Welker, Choromanski, Tombari, Purohit,
  Ryoo, Sindhwani, Lee, Vanhoucke, and Florence}]{Zeng2022SocraticMC}
Zeng, A.; Wong, A.~S.; Welker, S.; Choromanski, K.; Tombari, F.; Purohit, A.;
  Ryoo, M.~S.; Sindhwani, V.; Lee, J.; Vanhoucke, V.; and Florence, P.~R. 2022.
\newblock Socratic Models: Composing Zero-Shot Multimodal Reasoning with
  Language.
\newblock \emph{ArXiv}, abs/2204.00598.

\bibitem[{Zhang et~al.(2022{\natexlab{a}})Zhang, Zhang, Hu, Chen, Li, Dai,
  Wang, Yuan, Hwang, and Gao}]{glip2}
Zhang, H.; Zhang, P.; Hu, X.; Chen, Y.-C.; Li, L.~H.; Dai, X.; Wang, L.; Yuan,
  L.; Hwang, J.-N.; and Gao, J. 2022{\natexlab{a}}.
\newblock GLIPv2: Unifying Localization and Vision-Language Understanding.
\newblock \emph{ArXiv}, abs/2206.05836.

\bibitem[{Zhang et~al.(2021)Zhang, Guo, Zhang, Li, Miao, Cui, Qiao, Gao, and
  Li}]{pointclip}
Zhang, R.; Guo, Z.; Zhang, W.; Li, K.; Miao, X.; Cui, B.; Qiao, Y.~J.; Gao, P.;
  and Li, H. 2021.
\newblock PointCLIP: Point Cloud Understanding by CLIP.
\newblock \emph{2022 IEEE/CVF Conference on Computer Vision and Pattern
  Recognition (CVPR)}, 8542--8552.

\bibitem[{Zhang et~al.(2022{\natexlab{b}})Zhang, Roller, Goyal, Artetxe, Chen,
  Chen, Dewan, Diab, Li, Lin, Mihaylov, Ott, Shleifer, Shuster, Simig, Koura,
  Sridhar, Wang, and Zettlemoyer}]{Zhang2022OPTOP}
Zhang, S.; Roller, S.; Goyal, N.; Artetxe, M.; Chen, M.; Chen, S.; Dewan, C.;
  Diab, M.; Li, X.; Lin, X.~V.; Mihaylov, T.; Ott, M.; Shleifer, S.; Shuster,
  K.; Simig, D.; Koura, P.~S.; Sridhar, A.; Wang, T.; and Zettlemoyer, L.
  2022{\natexlab{b}}.
\newblock OPT: Open Pre-trained Transformer Language Models.
\newblock \emph{ArXiv}, abs/2205.01068.

\bibitem[{Zhou, Loy, and Dai(2021)}]{maskclip}
Zhou, C.; Loy, C.~C.; and Dai, B. 2021.
\newblock Extract Free Dense Labels from CLIP.
\newblock In \emph{European Conference on Computer Vision}.

\bibitem[{Zhou et~al.(2019)Zhou, Palangi, Zhang, Hu, Corso, and
  Gao}]{conimgcap2}
Zhou, L.; Palangi, H.; Zhang, L.; Hu, H.; Corso, J.~J.; and Gao, J. 2019.
\newblock Unified Vision-Language Pre-Training for Image Captioning and VQA.
\newblock \emph{ArXiv}, abs/1909.11059.

\bibitem[{Zhu et~al.(2023)Zhu, Chen, Shen, Li, and Elhoseiny}]{miniGPT4}
Zhu, D.; Chen, J.; Shen, X.; Li, X.; and Elhoseiny, M. 2023.
\newblock MiniGPT-4: Enhancing Vision-Language Understanding with Advanced
  Large Language Models.
\newblock \emph{ArXiv}, abs/2304.10592.

\end{thebibliography}

\clearpage

\setcounter{section}{0}

\begin{table*}[]
\centering
\small
\begin{tabular}{|l|c|c|c|c|c|}
\hline
& \# Trainable Parameter & \# Frozen Parameter & CC3M to MSCOCO(CIDEr) & MSCOCO(CIDEr) & Support Memory\\
\hline
CapDec & 181M & 0 & - & 0.918 & 0 \\
\hline
DeCap & 68M & 0 & 0.421 & 0.912 & 50,000 \\
\hline
Reproduced CapDec & 1.5M & 1.3B & 0.093 & 0.465 & 0 \\
\hline
Reproduced DeCap & 1.5M & 1.3B & 0.292 & 0.427 & 50,000 \\
\hline
MacCap & 5.7M & 1.3B &  0.525 & 0.687 & 0 \\
\hline
\end{tabular}
\label{tab:params}
\caption{The parameter size and captioning performance of baseline methods. We display zero-shot cross-domain and zero-shot in-domain captioning results from the main paper. The support memory is only used by DeCap, where 50,000 texts are used to construct a memory bank for inference.}
\end{table*}

\section{Detail of Baselines}
In this section, we provide a detailed explanation for the setting difference between our paper and previous methods DeCap\cite{DeCap}, CapDec\cite{CapDec}. The key difference is whether the language model or text decoder is trainable. DeCap and CapDec train the language model and achieve better performance in zero-shot in-domain captioning. However, we propose a more practical setting where the language model is frozen. The reason is that the large language model undergoes training through an auto-regression objective, thereby acquiring the capacity to proficiently execute a multitude of tasks, such as questing answering, translation, and automatic summarization. Training language model for a specific task leads to a degradation in other tasks, which is demonstrated in our zero-shot VQA experiments. Besides, the impressive performance of large language models comes with scale which makes finetuning language models computationally expensive. To provide a comprehensive picture, we collect the model parameters information in Table 1. We observe that the better performance of CapDec and DeCap in zero-shot in-domain captioning is achieved with more parameters, however, MacCap outperforms CapDec and DeCap in zero-shot cross-domain captioning with fewer parameters.

\section{Ablation Study on Hyperparameters}

\begin{figure}[htb]
  \centering
  \includegraphics[width=0.4\textwidth]{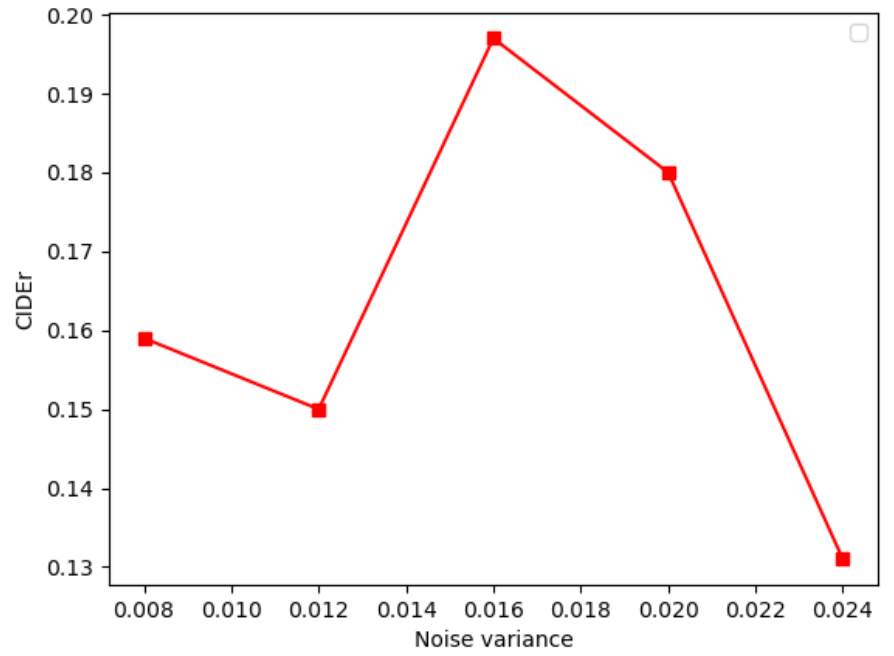}
   \caption{\textbf{Performance of MacCap with different training noise.} The MacCap is trained in CC3M dataset and tested on Flickr30K datasets.}
    \label{figure:noise}
    \vspace{-0.2cm}
\end{figure}

\textbf{Noise Variance}
The noise injection is utilized to reduce the widely observed \textit{modality gap} phenomenon in CLIP embedding space. We show the impact of noise variance when generating the text-region feature mentioned in Section 4.2 of the main paper. The results are shown in Figure~\ref{figure:noise}. We observed the best performance is achieved at noise variance 0.016 which is the same as suggested in \cite{CapDec}. Noise variances smaller than 0.016 cause performance drop due to the \textit{modality gap} while noise variances larger than 0.016 introduce extensive semantic ambiguity in text reconstruction training.

\begin{figure}[bt]
  \centering
  \includegraphics[width=0.4\textwidth]{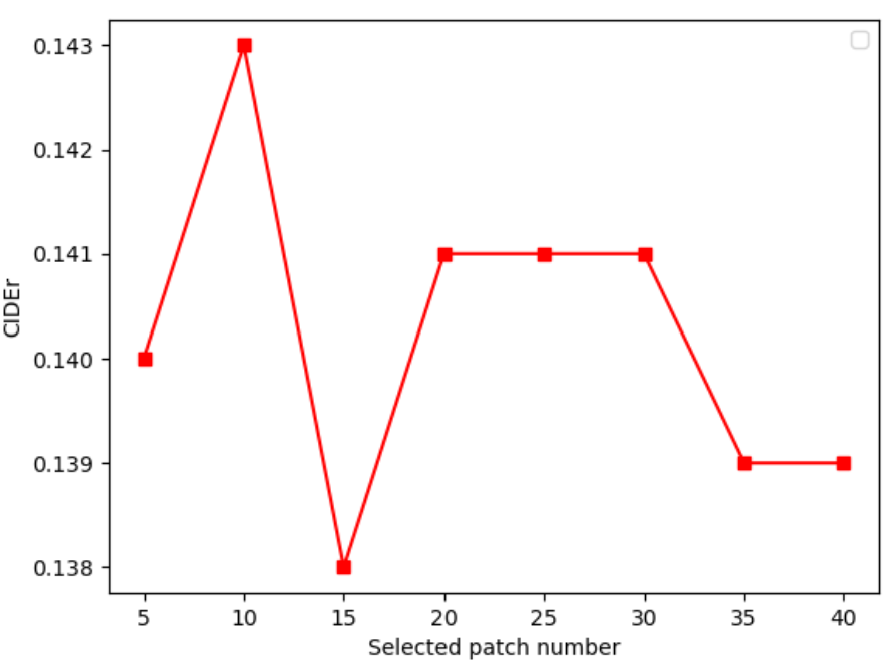}
   \caption{\textbf{Performance of MacCap with different patch numbers in inference.} The MacCap is trained in CC3M dataset and tested on Flickr30K datasets. The length of the text region feature in text reconstruction training is set to 10.}
    \label{figure:seq}
    \vspace{-0.2cm}
\end{figure}

\textbf{Number of Image Patches}
The sub-region feature aggregation in Section 4.3 of the main paper aims to extract the features of image subregions. We select image patches based on their attention scores as the patches with higher scores tend to contain more semantic information. We empirically study the number of selected patches, which indicates how much background information is introduced when generating captions. The results are shown in Figure~\ref{figure:seq}. We observe the best performance is achieved when the number of selected patches is the same as the number of noise-injected text-region features.

\section{Visualization of Image Sub-region}
\begin{figure}[tb]
  \centering
  \includegraphics[width=0.4\textwidth]{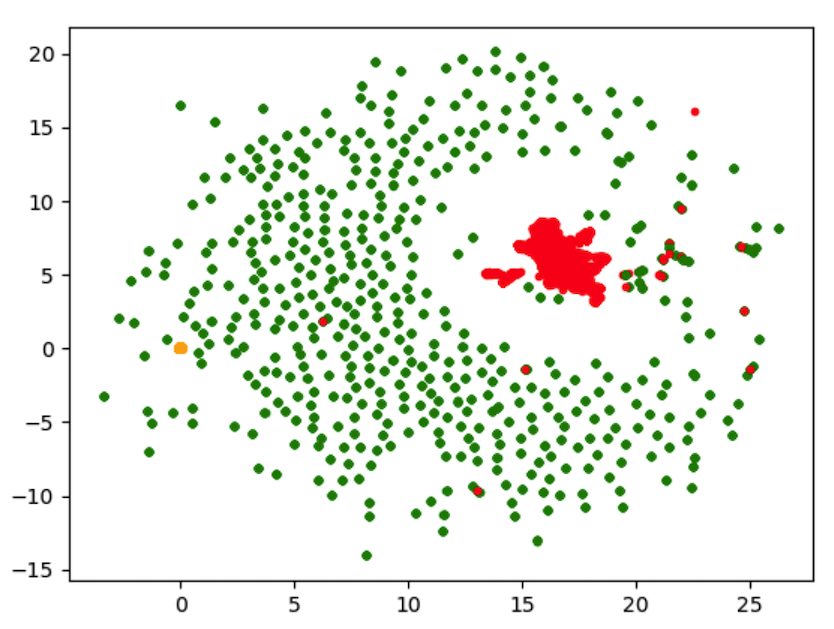}
   \caption{\textbf{Visualization of Embedding Distributions on MSCOCO.} The red points stand the global image embedding, each green point stand a local image embedding, and the yellow points near (0, 0) stand for the text embedding.}
    \label{figure:visulization}
    \vspace{-0.2cm}
\end{figure}

In this section, we visualize the distribution of the global image embeddings, local image embeddings (i.e. the embeddings of image sub-region), and the text embeddings. We show the relation between image subregions and corresponding captions in Section 3.1 of the main paper, which reveals that subregion embeddings may have a higher similarity with the caption text embedding as they can be the specific image regions described by the accompanying caption. We visualize 5000 samples from the MSCOCO dataset with the dimensionality reduction method UMAP following \cite{MindGap}. The results are shown in Figure~\ref{figure:visulization}. We observe the distances between text embeddings are relatively small and all of the text embeddings are distributed near the (0, 0). Furthermore, there exists a modality gap between global image embedding and text embedding. The local image embeddings are distributed in the surrounding region of global image embeddings and are relatively closer to text embeddings than the global ones.

\section{Details of Modality Gap Distribution Analysis} 
In this section, we provide implementation details of modality gap distribution experiments.

We have text modality representation $T^i \in \mathbb{R}^D$ and vision modality representation. To be noticed, for vision modality, we have two sets of representations: the global embedding representing the overall information and the patch embeddings representing the image subregions information. The global embedding is $I_c^i \in \mathbb{R}^D$, and the patch embedding is $I_p^i \in \mathbb{R}^{N \times D}$. Both global embedding and local embedding are obtained by CLIP. $D$ is the dimension of CLIP embedding and $N$ is the sequence length of CLIP. $i \in \{1, 2, ... , num\_samples\}$

For paired images and text descriptions, we evaluate the gap between text representation and both global and patch image representation. For each image-text pair, we compute $G^c_i = T^i - I_c^i$ and $G^p_i = repeat(T^i, N) - I_p^i$. $G^c_i \in \mathbb{R}^{D}$, $G^p_i \in \mathbb{R}^{N \times D}$.

We can check the overall distribution over all $D$ dimensions. In this case, we treat data in all dimensions equally. We compute the mean over all image-text pairs and draw the histogram of the data distribution. Thus we compute the mean of the gap distribution for global vision and language representation as $Avg_{(i, d)} (I_c^i[d])$, and the gap for local vision and language representation as $Avg_{(i, s, d)} (I_p^i[s][d])$. $s \in [N], d \in [D]$.

\end{document}